\documentclass{article}

\usepackage{PRIMEarxiv}

\usepackage[utf8]{inputenc} % allow utf-8 input
\usepackage[T1]{fontenc}    % use 8-bit T1 fonts
\usepackage{hyperref}       % hyperlinks
\usepackage{url}            % simple URL typesetting
\usepackage{booktabs}       % professional-quality tables
\usepackage{amsfonts}       % blackboard math symbols
\usepackage{fancyhdr}       % header
\usepackage{graphicx}       % graphics

\usepackage[numbers,sort&compress]{natbib}
\usepackage{amsmath}
\usepackage{amssymb}
\usepackage{amsthm}
\usepackage{caption}
\usepackage{enumitem}
\usepackage{xcolor}
\usepackage{geometry}
\usepackage{tikz}
\usetikzlibrary{shapes,arrows,positioning,calc,fit,shapes.geometric,arrows.meta}
\usepackage{listings}
\usepackage{upquote}
\usepackage{setspace}

\title{Pipeline for verifying LLM-generated mathematical solutions}

\author{
  Varvara Sazonova\footnotemark[1], Dmitri Shmelkin\footnotemark[2], Stanislav Kikot\footnotemark[3], Vasily Motolygin\footnotemark[4]
}

\begin{document}
\maketitle
\footnotetext[1]{Moscow State University, varvara26@yandex.ru}
\footnotetext[2]{Huawei Technologies Co., Ltd, Shmelkin.Dmitri@huawei.com}
\footnotetext[3]{Huawei Technologies Co., Ltd, kikot.stanislav@huawei.com}
\footnotetext[4]{motolygin98@gmail.com}

\begin{abstract}
With the growing popularity of Large Reasoning Models and their results in solving mathematical problems, it becomes crucial to measure their capabilities. We introduce a pipeline for both automatic and interactive verification as a more accurate alternative to only checking the answer which is currently the most popular approach for benchmarks. The pipeline can also be used as a generator of correct solutions both in formal and informal languages. 3 AI agents, which can be chosen for the benchmark accordingly, are included in the structure. The key idea is the use of prompts to obtain the solution in the specific form which allows for easier verification using proof assistants and possible use of small models ($\le 8B$). Experiments on several datasets suggest low probability of False Positives. The open-source implementation with instructions on setting up a server is available at ~\citep{lean4_verification_pipeline}
\end{abstract}

\section{Introduction}
With the growing popularity of Large Reasoning Models (that is, Large Language Models capable of reasoning which will be referred by LLMs later) and their results in solving mathematical problems, it becomes crucial to measure their capabilities. Most of the time, the evaluation is done only by checking the answer, however, this method is not entirely correct, as most human-oriented hard problems revolve not around providing the correct answer, which the LLMs are great at guessing, but around thorough reasoning. Even if the benchmark was specifically developed so that the current LLMs are unable to guess the answer without reasoning, as long as the correct answer is the only requirement for the solution, this behavior will continue to manifest in SOTA LLMs and problems with unknown answers or of niche subjects may become even more out of reach for them. The concerns of humans who graded LLM-generated solutions to 2025 International Mathematical Olympiad~\citep{imo} problems are shown in~\citep{imo-results}, with most of the solutions containing reasoning that was hard to follow. Thus it is important to be able to automatically assess reasoning quality of the solution.

We introduce a pipeline for that purpose, which includes two possible modes -- fully automatic, able to process multiple problems at once, and interactive or semi-automatic, which focuses on one problem and utilizes user feedback to overcome possible weak points of the implementation. A proof assistant is used to guarantee the correctness of the algorithm's judgement. In the provided implementation, Lean~\citep{lean4} is used as a proof assistant, however, switching the proof assistant is possible and will not change the steps of the algorithm. The output is a thorough report. If the solution is correct and verifiable by the pipeline, the report contains a full compiling Lean code correlating with the original solution in the natural language as well as details about each step. If the solution was incorrect, not thorough enough or not suitable for the current implementation, the report will include all finished steps and the errors so the reasons behind the algorithm stopping are clear.

The structure of the pipeline includes 3 LLMs: Solver LLM, Translator LLM and Prover LLM. Default models can be substituted for others similar to the role, making the pipeline adaptable to more benchmarks. The structure of the solution provided by the Solver LLM simplifies the processes of two other LLMs, improves the correlation between the original solution and Lean code and allows the use of smaller models. 

Our implementation is available at ~\citep{lean4_verification_pipeline}

\section{Related Work}

\subsection{Improvements of Informal LLM Reasoning}

Verification of LLM-generated solutions of mathematical problems, although important, is not a very popular area of research. Excluding the most similar area, Neural theorem proving, which will be mentioned below, there are no published works that can formally guarantee that the informal proof is correct, thus they cannot be considered verification fully. Works related to the improvements of LLM reasoning either use LLMs for verification~\citep{selfcheck} or are dataset-related: Zheng et al ~\citep{progressbench} provide a dataset as a benchmark for verification, the work of Zhang et al~\citep{deep-theorem} includes both the dataset and a specific RL (Reinforcement Learning) variant which are to improve quality of informal proving and the work of OpenAI~\citep{stepbystep} uses a human-made dataset in training to maximize the probability of the next generated line of the informal proof to be correct.

\subsection{Neural theorem proving}
Neural theorem proving is an area of research that employs models to aid in formal verification. Although similar, the goals of research in this area and LLM Verification are different. Results in Neural theorem proving are related to using a model in order to translate the mathematical text from an informal natural language (autoformalization) or to prove a theorem in the formal language, usually for the purpose of obtaining a correct proof from an agentic chain of LLMs. Our goal is to formally verify whether a given LLM-generated solution in the natural language is correct. Although the tools created for both subtasks (autoformalizing and proving) are insufficient to organize a required agentic chain on their own, they play a key role in the pipeline.

Autoformalization attempts, although unsuccessful, were made in Mizar~\citep{mizar} before the widespread use of AI models. Later they were resumed with the use of AI ~\citep{Wang_2020}. Most of the works in the area use the more popular Lean4 instead of Mizar. The availability of correct code examples in Lean4 and the library of formalized mathematical knowledge, Mathlib~\citep{mathlib}, is very useful for reinforcement learning and fine-tuning.
Since 2022, tools that use LLMs are being developed, such as LeanAide~\citep{leanaide} for formalization and Lean Copilot~\citep{leancopilot} for working with proofs in Lean. LLMs created or fine-tuned for the purposes of autoformalization and formal proof generation, like DeepSeek-Prover~\citep{xin2024deepseek},~\citep{huajian2024deepseekproverv15}, ~\citep{ren2025deepseek}, Goedel-Prover~\citep{{goedel-prover-v2}}, Kimina (Prover and Formalizer)~\citep{wang2025kimina}, AlphaProof~\citep{trinh2024alphaproof} and Aristotle~\citep{achim2025aristotleimolevelautomatedtheorem} were able to achieve great results, with the latter two winning the silver medal in the International Mathematical Olympiad, with many other models available publicly with a variation of weights. One of the last results of 2025 is G\" odel's Poetry~\citep{goedel-poetry} which employs earlier results and RAG for a structure with impressive results.

\subsection{Similar methodology}
With the creation of ai tools for autoformalization and proof generation, new methods had to be developed in order to determine the best strategy for an agentic chain. Kimina, Aristotle, G\" odel's Poetry and other ai models use their own methodologies to overcome difficulties related to harder problems, although one of the methods is used in almost all, if not all, pipelines: division of the solution into smaller statements -- lemmas, each of which is easier to consider on its own rather than the whole proof. One of the first instances of this strategy and the closest to our implementation is Draft, Sketch, Prove~\citep{draft-sketch-prove}, which proposes creating an easily formalized proof sketch, which will then be completed by the LLM. The method described as in the paper is suited for the Neural theorem proving and not so for LLM Verification, however, the idea of a sketch filled out with the LLM is present in our implementation with a reworking.

\section{Methods}

\subsection{Main Idea}\label{sub:main-idea}
As mentioned before, the problem of verification using LLMs and proof assistants includes two subtasks: autoformalizing the solution into Lean4 and using the prover LLM to complete the code. It is of utmost importance to not only minimize the amount of mistakes during the autoformalization, but to make sure that every important part of the solution is translated properly, furthermore, there should be a straightforward way to link all parts of the original sketch iff the original solution is correct. 

The key idea is to demand a specific solution from Solver LLM. The properties that we aim to satisfy in the original solution with the use of prompt engineering is as follows: 

A correct and sufficiently explained proof must have a following structure:
\begin{itemize}
	\item Each logical step of the proof (lemma) should be written in propositional logic as a set of premises and a conclusion. 
	\item The premises are either: 1) conclusions of previous logical steps 2) given in the statement 3) consist of well-known (Pythagoras theorem) or obvious (2 + 2 = 4) facts.
	\item Each logical step must have only one statement in the conclusion (without $\land$ or "if else" construction)
	\item Each logical step is correct and can be proven by a human fairly easily (for example, in no more than 3-5 completely formalized steps, where each step is axiomatically correct). 
\end{itemize}

The proof with m lemmas will have the following form:
\[
\left\{
\begin{array}{@{}l@{}}
A_1 \land \dots \land A_n \to C \\[4pt]
\bigl\{ F_{1,1} \land \dots \land F_{1,i_1} \bigr\} 
    \land \bigl\{ A_{j_{1,1}} \land \dots \land A_{j_{1,k_1}} \bigr\} 
    \to L_1 \\[4pt]
\bigl\{ F_{2,1} \land \dots \land F_{2,i_2} \bigr\} 
    \land \bigl\{ A_{j_{2,1}} \land \dots \land A_{j_{2,k_2}} \bigr\} 
    \land \bigl\{ L_1 \bigr\} 
    \to L_2 \\[4pt]
\bigl\{ F_{3,1} \land \dots \land F_{3,i_3} \bigr\} 
    \land \bigl\{ A_{j_{3,1}} \land \dots \land A_{j_{3,k_3}} \bigr\} 
    \land \bigl\{ L_1 \land L_2 \bigr\} 
    \to L_3 \\[4pt]
\vdots \\[4pt]
\bigl\{ F_{m,1} \land \dots \land F_{m,i_m} \bigr\} 
    \land \bigl\{ A_{j_{m,1}} \land \dots \land A_{j_{m,k_m}} \bigr\} 
    \land \bigl\{ L_1 \land \dots \land L_{m-1} \bigr\} 
    \to C
\end{array}
\right.
\]
Where $A_1, ..., A_n$ are premises given in the problem statement and each premise may include quantifiers, C is the statement that needs to be proved, $F_{i,j}$ are facts that are easy to verify, $A_{j_{p,q}} \in \{A_1, \dots, A_n\}.$, $L_i$ is the conclusion of lemma i and each $\bigl\{ ... \bigr\}$ segment is either empty or contains only elements of given structure.

This structure not only provides clear implications for each lemma, but reduces error probability and makes the translation, proving each lemma and linking all the lemmas together easier for a user to verify in interactive mode and for the other AI agents to perform. The pipeline is most suited to the solutions with such structure.

\subsection{Agentic Chain}
As mentioned in the introduction, there are 3 AI models used in the pipeline, each having their own role: Solver LLM, Translator LLM and Prover LLM. The $\le 8B$ restriction mentioned below is non-mandatory for the pipeline, moreover, larger models may improve Recall as long as they are suited for the role in the pipeline and the dataset. Each LLM may be switched, however, since the pipeline requires prompt engineering and scripts that are specific to the behavior of each model, we have decided to assess and improve our results on fixed LLMs mentioned below.

Solver LLM is a LLM that needs to be assessed, its role is to generate the original solution in the natural language. Each model would need its own prompts in order to get the solution in the structure mentioned earlier, with some models being more resistant to the prompt. If the aforementioned structure is not obtained properly for a specific problem, it is highly likely to not be able to be verified by the pipeline (resulting in a False Negative), therefore the most useful models for this role should have more potential of being influenced by the prompts. In our experiments, out of publicly available $\le 8B$ models, we find that Qwen~\citep{qwen2025} distillations are the easiest to work with, that is why we use Qwen3-8B~\citep{qwen3-8b} in our experiments.

Translator LLM is a model created specifically for the role of autoformalization. Out of publicly available $\le 8B$ models, we use Kimina-Autoformalizer-7B~\citep{kimina-auto-7b}. Kimina is trained with the use of RL, hence Kimina-Autormalizer is not as responsive to prompts (that is, the quality may sharply decrease and the amount of hallucinations may increase when the prompt differs significantly from the developers' version), which we overcome with scripts related to specific situations where mistakes are systematically made. Theoretically, the best formalizer model would be the one fine-tuned for lemmas with such specific structure and the dataset in mind, however, this requires a lot of resources, thus the scripts are a more beginner-friendly option.

Translator LLM is a model created specifically for the role of proving theorems in Lean, namely, facts and lemmas in our pipeline. Out of publicly available $\le 8B$ models, we use Kimina-Prover-Preview-Distill-7B~\citep{kimina-prover-7b} as well as several scripts.

\subsection{Algorithm Steps}
\begin{figure}[htbp]
  \centering
  \includegraphics[width=1\textwidth]{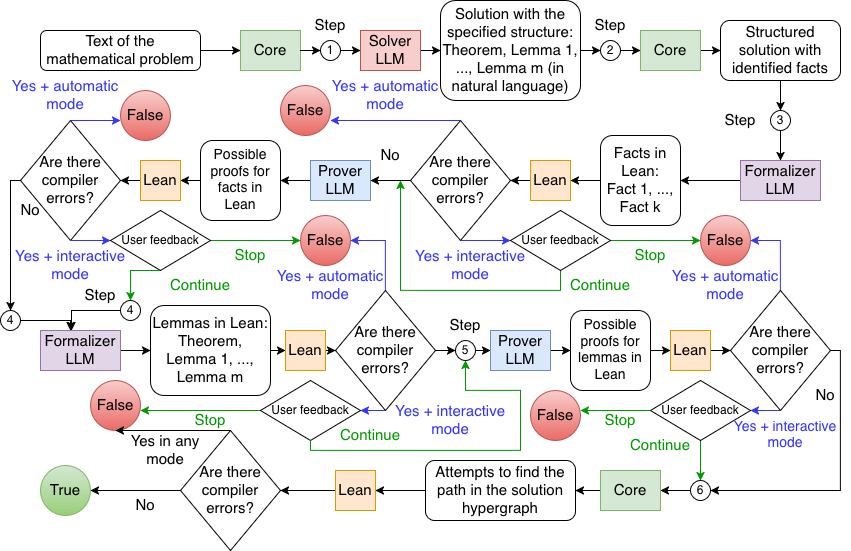}
  \caption{Scheme of the algorithm}
  \label{fig:scheme}
\end{figure}
The algorithm for both automatic and interactive modes consists of 6 steps: getting a specifically structured solution, analyzing its structure with the script, formalizing all "facts" and checking their evidence, formalizing all lemmas and checking the quality of the translation with the script, proving all lemmas and linking the whole proof. Figure ~\ref{fig:scheme} displays the scheme of the algorithm.

The full description of each step is as follows:

\begin{enumerate}
	\item Getting a specifically structured solution: This step is the same for both modes -- the Solver LLM is sent a specific prompt for the structure of the solution and the original problem statement. The output is the solution of the form described in section \ref{sub:main-idea}. If the solution severely deviates from the form, unless a new prompt is developed, the problem cannot be verified in the pipeline. If the deviations are small, they will be removed in the next step. 
	
	It should be noted that the prompts may be reworked in order to properly assess different types of problems. For example, solutions for word-based problems should include a fixed set variables, each having a fixed type, otherwise, difficulties arise on later steps.
	
	\item Analyzing the structure of the solution with the script: This step is the same for both modes -- Firstly, the solution is checked to be of proper form (excluding the criterion about correctness of each lemma which will be checked on later steps) and rewritten for easier translation. For example, if present, tautologies will be removed and if the lemma's conclusion is of form B $\land$ C, it will be rewritten as two lemmas. The other changes are mostly related to Translator LLM, that is, reworking statements that are highly probable to cause errors. Secondly, the "facts", that is, premises that were not present in the problem statement or previous lemmas, are found.
	\item Formalizing all "facts" and checking their evidence: Each "fact" from the second step is firstly formalized by the Translator LLM. In automatic mode, if the translation has errors and unable to compile in Lean4, the algorithm will stop with the negative answer (that is, the solution is incorrect or unable to be checked), in interactive mode, the user will have the option to either stop the algorithm with the negative answer or provide a correct translation. 
	
	Once the formalization has no errors, Prover LLM will try to prove it. It should be noted that sometimes the fact is true in the context of the problem, but was rewritten in a non-obvious way. Thus, context from earlier lemmas is added as premises. In automatic mode, if the proof is unable to compile, the algorithm will stop with the negative answer, in interactive mode, the user will have the option to either: a) continue without this fact b) output that this fact is true and for it to be accepted without proof (with the use of Lean4's tactic $sorry$ which allows the theorem without proof to compile) c) output that this fact is false and stop the algorithm with the negative answer d) retry, giving Prover LLM another opportunity to prove the fact e) output that the fact was translated incorrectly and provide the correct translation, after which the Prover LLM restarts.
	\item Formalizing all lemmas and the theorem and checking the quality of the translation with the script: similarly to step 3, each lemma is formalized by Translator LLM. In automatic mode, if the translation has errors and unable to compile in Lean4, the algorithm will stop with the negative answer, in interactive mode, the user will have the option to either stop the algorithm with the negative answer or provide a correct translation. The formalizations are then checked with the script to ensure that no important information from the original solution is missing or replaced and the structure remains, as well as that all the types in Lean are chosen correctly.
	\item Proving all lemmas: similarly to step 3, each formalized lemma is attempted to be proven by the Prover LLM with added context from previous lemmas. In automatic mode, if the proof is unable to compile, the algorithm will stop with the negative answer, in interactive mode, the user will have the same options as in step 3.
	\item Linking the whole proof: This step is the same for both modes -- after all lemmas have been proven, the scripts link them together using tactics which comes down to finding the path in the hypergraph of the solution. If the structure remains, this path will be found using multiple logical rules of inference, which corresponds to the tactic solve\_by\_elim in Lean. After the last step, the report of all the steps of the algorithm is compiled.
\end{enumerate}

\subsection{Introduction of Variables}\label{sub:intro-var}
While working on the pipeline, our team has noticed the problems with verifying text-based problems: due to each lemma being considered on its own, if the variables were not explicitly stated (as in '$3*x = 9 \to x = 3$') or their type was not obvious in the lemma (for example, the problem might mention that an integer amount of cakes was eaten, but this would not be stated in the lemma, just used without 'formal' justification), this would often lead to difficulties with properly formalizing the lemmas or finding the correct path in the solution hypergraph if lemmas were proven for wrong variables with wrong types. That is why, in both interactive and automatic modes, there is an option to disable or enable $introduction$ $of$ $variables$ - a specific prompt that demands that the Solver LLM highlights the variables and their types that were either given in problem statement or might be introduced in the solution) and uses them and only them in the solution with the specific type. This option is not always optimal and the best results are achieved when using the combination of introducing and not introducing the variables.

\subsection{Example Of Pipeline Verification}
\begin{figure}[htbp]
  \centering
  \includegraphics[width=1\textwidth]{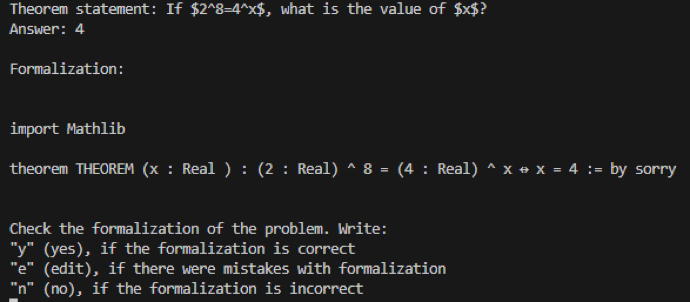}
  \includegraphics[width=1\textwidth]{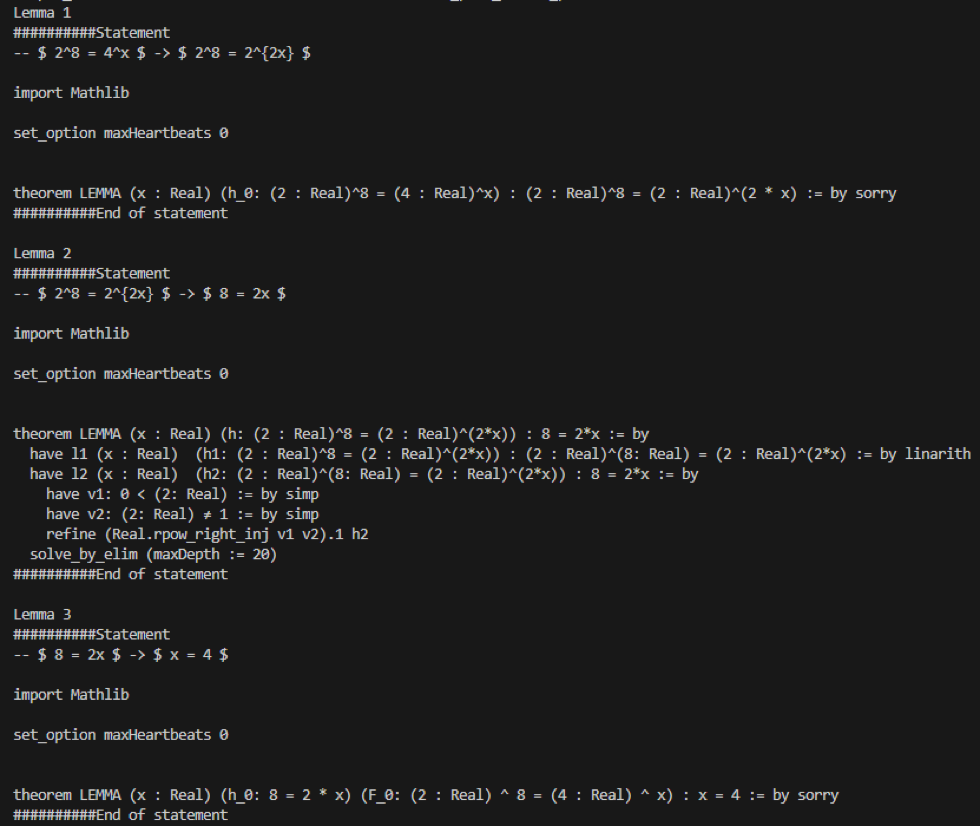}
  \caption{Verification of an easy problem, part 1}
  \label{fig:example1}
\end{figure}
\begin{figure}[htbp]
  \centering
  \includegraphics[width=1\textwidth]{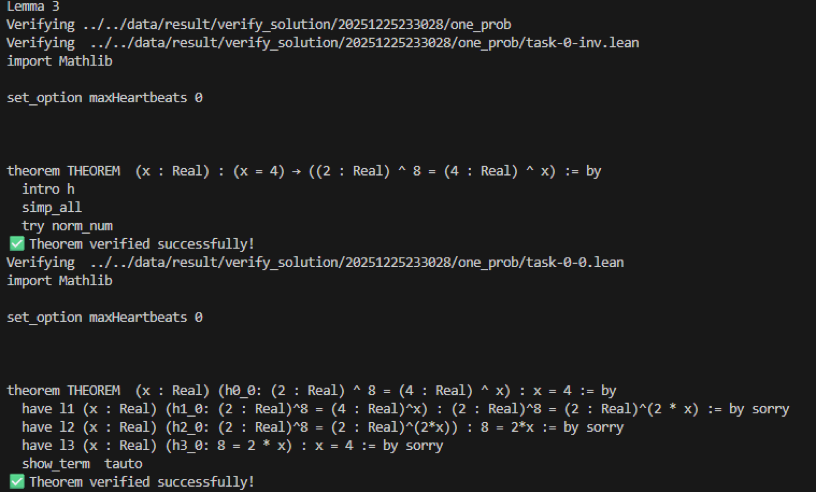}
  \caption{Verification of an easy problem, part2}
  \label{fig:example2}
\end{figure}
\begin{figure}[htbp]
  \centering
  \includegraphics[width=1\textwidth]{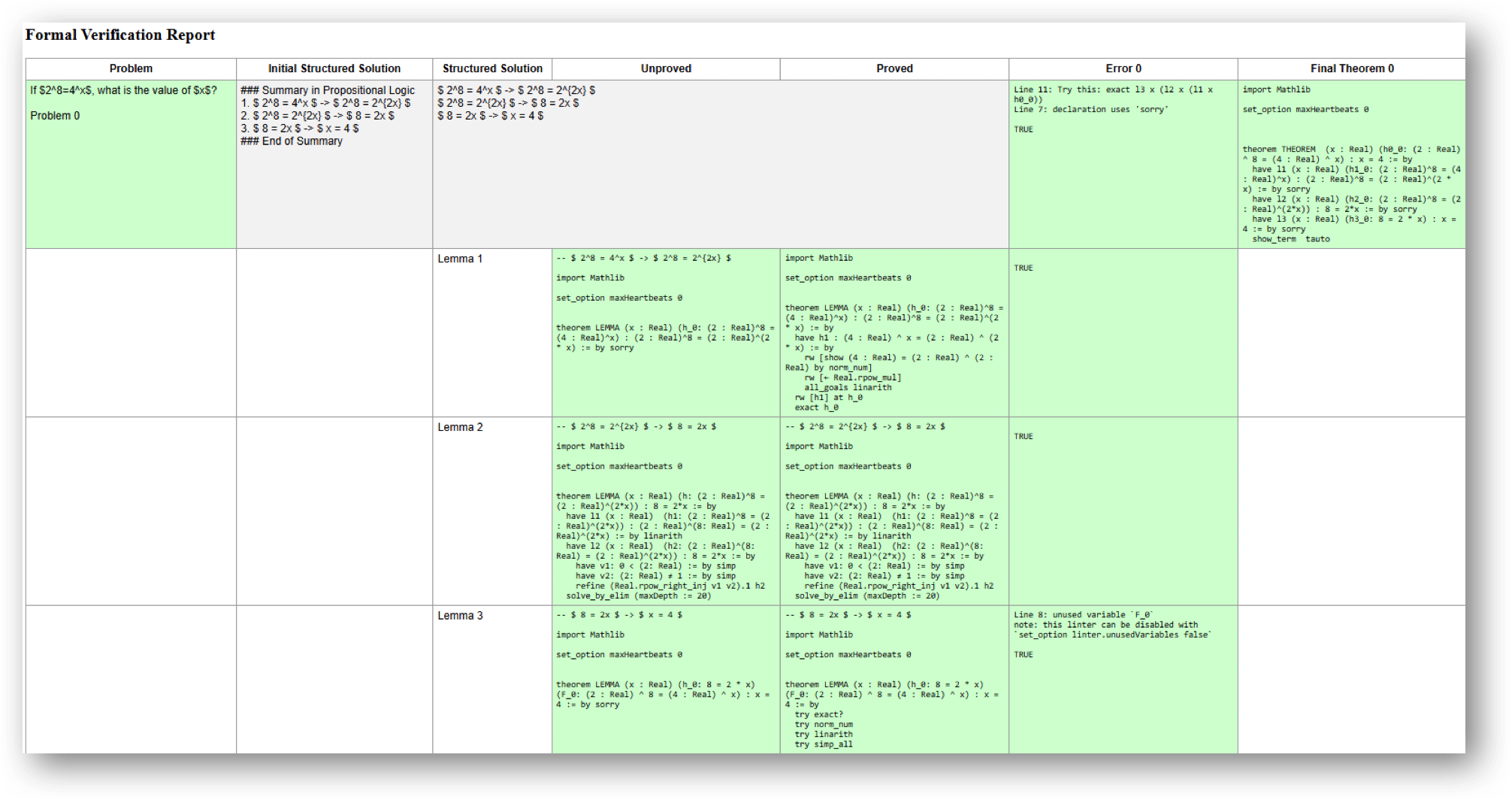}
  \caption{Report of the verification}
  \label{fig:report}
\end{figure}

Figures ~\ref{fig:example1} and ~\ref{fig:example2} showcase the work of the algorithm in our implementation at ~\citep{lean4_verification_pipeline} on the easy problem during the interactive mode, which finished without any need of human feedback. Figure ~\ref{fig:report} shows the report created for that instance of verification.

\section{Results}

\subsection{Expected Behavior In Possible Cases}
\begin{figure}[htbp]
  \centering
  \includegraphics[width=1\textwidth]{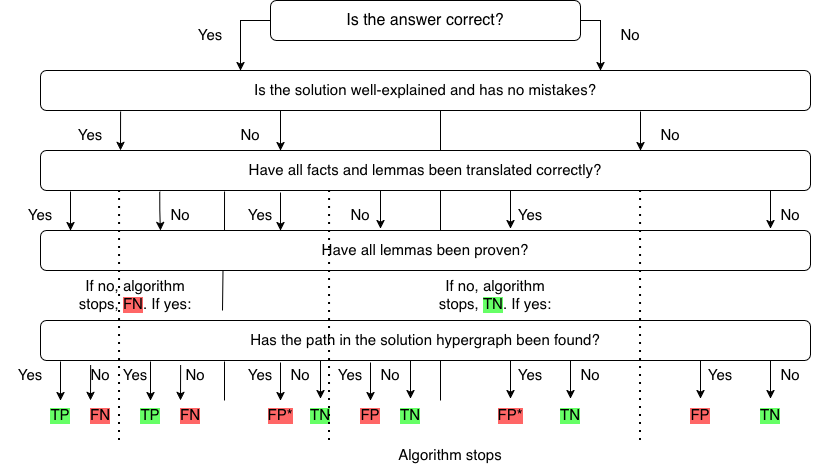}
  \caption{Diagram of cases}
  \label{fig:diagram}
\end{figure}

Figure ~\ref{fig:diagram} displays the formation of the error matrix. 

Correctness of the answer is usually easy to check if the benchmark provides the ground truth. Verifying whether there are mistakes in the solution is also straightforward, however, whether the solution is well-explained is not always understandable even by mathematicians when grading human-made solutions. 

For example, some competitions may grade the solution without proper reference to theorems (say, Ceva's theorem) as correct while others penalize the solution if the referenced fact was unproven in it. If the human is grading the work, they may intuitively understand logical steps and transitions, but still penalize the author for not being thorough enough, and it is almost impossible to formalize the blurred limits of what is accepted as proper reasoning. 

At this point of development, we consider the solution to be well-explained iff each lemma is able to be proven with earlier context in automatic mode (or however the user decides in the interactive mode) and all the lemmas are able to be linked together either without any assistance or with the use of several scripts (for example, if the last lemma concludes not with the expected theorem statement, but the statement follows from that conclusion, we can implement the last lemma and check using Prover LLM whether that is true and finish the proof automatically. Thus, this solution is considered well-explained). 

It has been experimentally proven to be highly unlikely to encounter a False Positive in both modes, excluding the following scenario: sometimes the problem is too easy (for example, 'find the GCD of 3 numbers'), the Solver LLM becomes confused and is unable to provide the adequate structure for the solution, however, the theorem in Lean4 may be able to be proven even without the structure due to the optimization of the proof assistant. In other cases, either the solve\_by\_elim tactic will not work or the translated lemma will not be able to be proven. Cases where the wrong solution for a hard enough problem was able to be translated into a correct proof have not been noticed experimentally, not to mention that the translation is checked with the script in order to catch possible inaccuracies or hallucinations. To further minimize the probability of False Positives appearing, one can provide the correct formalization of the main theorem instead of the Translator LLM. 

Thus, excluding the aforementioned class of problems, one can attempt to generate a plethora of correct solutions. If the problem statement was translated correctly (or the correct translation was given) and Lean code compiles, the solution in the formal language is correct (which can be obtained automatically) and the structure of the translated (as well as the original) solution shows explicitly what (correct) logical steps have been taken.

\subsection{Benchmarks For Evaluation}
Unfortunately, not all classes of mathematical problems are currently able to be processed by the pipeline. The list of problems that are not able to be verified includes, but not limited to: a) all geometrical problems due to the difficulties of formalizing geometrical objects and proofs in Lean4, b) most problems that revolve around higher mathematics due to proper theorems and concepts needed for the proof not being present in Lean4, c) all of the problems that are unable to be solved with the respect to the required structure by Solver LLM.

 In our experiments, the Solver was unable to provide well-explained proofs for many olympiad problems, which is why we chose Math-500~\citep{math500} and its subsets as the main datasets for the experiments, namely:
 \begin{enumerate}
	\item The whole Math-500: the whole dataset contains many problems that are unable to be proven correctly in the current developmental phase (namely combinatorics-related problems), which is why it was only used to evaluate FP and Precision.
	\item "easy": a dataset consisting of 10 problems from Math-500, the hardest of them being solved by 5 easy lemmas. Mostly used to check FPs in "easy to the point where Lean4 can solve it on its own" problems and as an autotest for the new implementations of the pipeline.
	\item "similar": the main dataset of 150 problems from Math-500, excluding most of the problems that are nearly impossible to formalize or prove. 
\end{enumerate}

The datasets are available at ~\citep{lean4_verification_pipeline} with the implementation.

\subsection{Benchmark Results} 
For interactive mode, it is possible to achieve 0 FN and 0 FP, depending on the user's knowledge of Lean. For automatic mode, the results are as follows:
In "easy" dataset:
 \begin{enumerate}
	\item Accuracy = $0.840 \pm 0.102$
	\item Precision = $0.822 \pm 0.113$
	\item Recall = $0.953 \pm 0.058$
\end{enumerate}

In "similar" dataset:
 \begin{enumerate}
	\item Accuracy = $0.847 \pm 0.016$
	\item Precision = $0.984 \pm 0.008$ (excluding the "easy" problems makes precision almost 1)
	\item Recall = $0.609 \pm 0.006$ (can be improved with scripts and larger models)
\end{enumerate}

In the whole Math-500 dataset Precision = $0.942 \pm 0.008$.

\section{Discussion}

\subsection{Comparison With Non-Verification Tools}
In the beginning of development, the following experiment was conducted: a dataset of 200 solutions was considered
 \begin{itemize}
	\item 110 of which were original LLM-generated solutions for problems from MATH-500 subset of easy non-geometric tasks
	\item Out of these 110 solutions, 50 are correct and well-explained while 60 are incorrect: 35 containing mistakes and 25 being not well explained 
	\item The rest 90 of which were the following: 50 almost correct solutions that were poisoned with some small wrong statements, 10 solutions consisting of just of the correct answer without any explanations, 30 cases the solution was originally correct but the problem statement was exchanged. 
\end{itemize}

We ran 200 problems through Qwen, asking it whether the solution is correct. As for our pipeline, since the problems were very easy (but not to the point of providing risk of False Positives), they didn't need to have the aforementioned structure in order to be understood by the formalizer, and splitting the solution into lemmas (without changing anything in the text) was the only preprocessing done by our team.

Qwen was able to show that all 50 correct solutions are correct, however, it had 85 False Positives. Our tool approved 43 correct solutions due to difficulties with formalization (and all 50 with user feedback) and evaluated as incorrect all incorrect.

After but still in the early stages in development, this pipeline was compared with the pipeline consisting of only Kimina-Autoformalizer + Prover with the latter having a natural language solution as a hint in the prompt (formalizing the problem statement using Kimina-Autoformalizer and giving the whole statement to Kimina-Prover rather than lemmas in our original pipeline). The goal of this pipeline was to check not solutions but answers. On easy, the results were relatively the same (at least 9 out of 10 problems were able to be solved each time), but on similar, the other pipeline was able to check on average 7 more answers. This does not diminish our results, as the goals of the pipelines are different and the amount of true negatives is not the same, although this does show that the structures developed in relation to Neural Theorem Proving are suited better for the purposes of that area.

\subsection{Bottlenecks And Limitations}
One of the largest bottlenecks of the area of research is that mathematical accuracy for proof assistants and for humans is not the same, the former being incredibly stricter. Many of the False Negatives arise due to type problems or small missed steps that are normal for a human but fatal for a proof assistant. This seems possible to overcome only to some extent and with proper ML and scripts for systematic difficulties.

Since the pipeline includes an agentic chain, if one agent in the chain makes a mistake, the whole process becomes incorrect, however, automatic verification seems impossible without agents at the moment, thus there is nothing that can be fixed.

An important task to consider is the need to not lose context between each lemma and to consider various non-equivalent notations in Lean4. This is mostly done through scripts, since giving too much context in a prompt can cause the agent to hallucinate, and the scripts require more work.

The pipeline is harder to generalize onto harder (more olympiad-oriented problems or if they belong to specific niche domains), since each step would be harder to formalize as an “easy/evident” lemma, which would become even harder for huge projects with a lot of dependencies. This might require another methodology.

\subsection{Open Source Implementation and Server Setup}
The open-source implementation with instructions on setting up a server is available at ~\citep{lean4_verification_pipeline}. There are 2 branches - main (automatic) and interactive. The setup instructions available at README.md are the same for both branches. The repository includes:
 \begin{itemize}
	\item Two modes, each having an option to introduce/not introduce variables (see section ~\ref{sub:intro-var})
	\item Datasets that were used for results
	\item The pipeline for answer verification mentioned at the beginning of the section
	\item The option for easier A/B testing of new features with a custom report, which provides the comprehensive comparison between earlier and new results on a fixed dataset.
 \end{itemize}

\section{Conclusion}
This paper introduces a pipeline for verification of LLM-generated mathematical problems. The structure has an automatic and interactive mode and can be adapted to a specific benchmark by changing the AI models in the agentic chain. We highlight the complex relationship between mathematical accuracy for humans and proof assistants as well as various bottlenecks and ways to improve the algorithm. The provided architecture is highly improbable to produce false positive results, excluding a class of very easy problems, which makes it suitable for generation of correct solutions. The pipeline is publicly available with a variety of features and an option to change each LLM agent in order to improve results. We plan to further improve the algorithm for a plethora of problems with varying difficulty with scripts and new theoretical ideas and methodologies in order to achieve better verification capabilities.

\bibliographystyle{plainnat}

\begin{thebibliography}{99}

\bibitem{lean4_verification_pipeline}
Varvara Sazonova, Dmitri Shmelkin, Stanislav Kikot and Vasily Motolygin.
Open source implementation of the pipeline described in the paper.
\url{https://github.com/LogicEnj/lean4_verification_pipeline}

\bibitem{imo}
\url{https://www.imo-official.org/}

\bibitem{imo-results}
\url{https://matharena.ai/imo/}

\bibitem{lean4}
Leonardo de Moura and Sebastian Ullrich.
The Lean 4 Theorem Prover and Programming Language.
In \textit{International Conference on Automated Deduction (CADE)}, 2021.

\bibitem{selfcheck}
Ning Miao, Yee Whye Teh, Tom Rainforth
Selfcheck: Using llms to zero-shot check their own step-by-step reasoning 
\textit{arXiv preprint arXiv:2308.00436}, 2023.

\bibitem{progressbench}
Zheng, Chujie, et al.
Processbench: Identifying process errors in mathematical reasoning 
In \textit{Proceedings of the 63rd Annual Meeting of the Association for Computational Linguistics (Volume 1: Long Papers)}, 2025.

\bibitem{deep-theorem}
Zhang Z. et al. 
Deeptheorem: Advancing llm reasoning for theorem proving through natural language and reinforcement learning 
\textit{arXiv preprint arXiv:2505.23754}, 2025.

\bibitem{stepbystep}
Lightman, H., Kosaraju, V., Burda, Y. et al.
Let's verify step by step. 
In \textit{The Twelfth International Conference on Learning Representations.}, 2023.

\bibitem{mizar}
\url{https://mizar.uwb.edu.pl/}

\bibitem{Wang_2020}
Wang Q. et al. 
Exploration of neural machine translation in autoformalization of mathematics in Mizar.
In \textit{Proceedings of the 9th ACM SIGPLAN International Conference on Certified Programs and Proofs}, 2020.

\bibitem{mathlib}
The mathlib Community.
The Lean Mathematical Library.
In \textit{CPP}, 2020.

\bibitem{leanaide}
Agrawal, Ayush, et al. 
"Towards a mathematics formalisation assistant using large language models." 
\textit{arXiv preprint arXiv:2211.07524}, 2022.

\bibitem{leancopilot}
Song P., Yang K., Anandkumar A. 
Lean copilot: Large language models as copilots for theorem proving in lean
\textit{arXiv preprint arXiv:2404.12534}, 2024.

\bibitem{xin2024deepseek}
Huajian Xin, Daya Guo, et al.
DeepSeek-Prover: Advancing Theorem Proving in LLMs through Large-Scale Synthetic Data.
\textit{arXiv preprint arXiv:2405.14333}, 2024.

\bibitem{huajian2024deepseekproverv15}
Xin Huajian, Z. Z. Ren, et al.
DeepSeek-Prover-V1.5: Harnessing Proof Assistant Feedback for Reinforcement Learning and Monte-Carlo Tree Search.
\textit{arXiv preprint arXiv:2408.08152}, 2024.

\bibitem{ren2025deepseek}
Z. Z. Ren, Zhihong Shao, et al.
DeepSeek-Prover-V2: Advancing Formal Mathematical Reasoning via Reinforcement Learning for Subgoal Decomposition.
\textit{arXiv preprint arXiv:2504.21801}, 2025.

\bibitem{goedel-prover-v2}
Yong Lin, Shange Tang, Bohan Lyu, et al.
Goedel-Prover-V2: Scaling Formal Theorem Proving with Scaffolded Data Synthesis and Self-Correction.
\textit{arXiv preprint arXiv:2508.03613}, 2025.

\bibitem{wang2025kimina}
Haiming Wang, Mert Unsal, et al.
Kimina-Prover Preview: Towards Large Formal Reasoning Models with Reinforcement Learning.
\textit{arXiv preprint arXiv:2504.11354}, 2025.

\bibitem{trinh2024alphaproof}
Trieu H. Trinh, Yuhuai Wu, et al.
Solving Olympiad Geometry without Human Demonstrations.
\textit{Nature}, 2024.

\bibitem{achim2025aristotleimolevelautomatedtheorem}
Tudor Achim, Alex Best, et al.
Aristotle: IMO-level Automated Theorem Proving.
\textit{arXiv preprint arXiv:2510.01346}, 2025.

\bibitem{goedel-poetry}
Davis, Kelly J. 
G\" odel's Poetry.
\textit{arXiv preprint arXiv:2512.14252}, 2025.

\bibitem{draft-sketch-prove}
Jiang, Albert Q., et al. 
Draft, sketch, and prove: Guiding formal theorem provers with informal proofs.
\textit{arXiv preprint arXiv:2210.12283 (2022).}

\bibitem{qwen2025}
Qwen.
Qwen3: Think Deeper, Act Faster.
\url{https://qwen.ai/blog?id=1e3fa5c2d4662af2855586055ad037ed9e555125&from=research.research-list}, 2025.

\bibitem{qwen3-8b}
\url{https://huggingface.co/Qwen/Qwen3-8B}

\bibitem{kimina-auto-7b}
\url{https://huggingface.co/AI-MO/Kimina-Autoformalizer-7B}

\bibitem{kimina-prover-7b}
\url{https://huggingface.co/AI-MO/Kimina-Prover-Preview-Distill-7B}

\bibitem{math500}
OpenAI.
Math-500.
\url{https://huggingface.co/datasets/HuggingFaceH4/MATH-500}

\end{thebibliography}

\end{document}